\documentclass[runningheads]{llncs}
\usepackage{adjustbox}
\usepackage[colorinlistoftodos]{todonotes}
\usepackage[colorlinks=true, allcolors=blue]{hyperref}
\usepackage{graphicx}
\usepackage{cite} 
\usepackage{times}
\usepackage{epsfig}
\usepackage{amsmath}
\usepackage{amssymb}
\usepackage{mwe}
\usepackage{acro}
\usepackage{amssymb}
\usepackage{xcolor,colortbl}
\usepackage{tabularx}
\usepackage{relsize}
\usepackage{pifont}
\usepackage{booktabs} 
\usepackage{multirow}
\usepackage{multicol}
\usepackage{adjustbox}
\usepackage{amssymb}
\usepackage{float}
\usepackage{epsfig}
\usepackage{amsmath}
\usepackage{amssymb}
\usepackage{algorithm}
\usepackage{algorithmic}
\usepackage{subfigure}
\usepackage{caption}
\usepackage{epsfig}
\usepackage{epstopdf}
\usepackage{wrapfig}
\usepackage{multirow}
\usepackage{array}
\usepackage{tabulary}
\usepackage{graphicx}
\usepackage{color}
\usepackage{marvosym}

\begin{document}
\title{$M^{2}$Fusion: Bayesian-based Multimodal Multi-level Fusion on Colorectal Cancer Microsatellite Instability Prediction}
% \author{Submission 1} 
\author{Quan Liu\inst{1,2} \thanks{Work was done during an internship at Alibaba DAMO Academy. Corresponding author$^{(\textrm{\Letter})}$:\href{mailto:zyliu@163.com}{zyliu@163.com}}  \and
Jiawen Yao\inst{1,3}\and
Lisha Yao\inst{4,5} \and
Xin Chen\inst{4} \and
Jingren Zhou\inst{1} \and
Le Lu\inst{1} \and
Ling Zhang\inst{1} \and
Zaiyi Liu\inst{4}$^{(\textrm{\Letter})}$ \and
Yuankai Huo\inst{2}}

\authorrunning{Q. Liu et al.}

\institute{DAMO Academy, Alibaba Group
\and 
Vanderbilt University, Nashville TN, USA \and Hupan Lab, 310023, Hangzhou, China
 \and
Guangdong Provincial People’s Hospital, Guangzhou, China \and
South China University of Technology, Guangzhou, China
}

%
% \authorrunning{H. Yang et al.}
% \author{submission 569}
% \institute{******}
% \institute{
% *********
% }
% \email{yuankai.huo@vanderbilt.edu}
%
\maketitle              % typeset the header of the contribution
\begin{abstract}
Colorectal cancer (CRC) micro-satellite instability (MSI) prediction on histopathology images is a challenging weakly supervised learning task that involves multi-instance learning on gigapixel images. To date, radiology images have proven to have CRC MSI information and efficient patient imaging techniques. Different data modalities integration offers the opportunity to increase the accuracy and robustness of MSI prediction. Despite the progress in representation learning from the whole slide images (WSI) and exploring the potential of making use of radiology data, CRC MSI prediction remains a challenge to fuse the information from multiple data modalities (e.g., pathology WSI and radiology CT image). In this paper, we propose $M^{2}$Fusion: a Bayesian-based multimodal multi-level fusion pipeline for CRC MSI. The proposed fusion model $M^{2}$Fusion is capable of discovering more novel patterns within and across modalities that are beneficial for predicting MSI than using a single modality alone, as well as other fusion methods. The contribution of the paper is three-fold: (1) $M^{2}$Fusion is the first pipeline of multi-level fusion on pathology WSI and 3D radiology CT image for MSI prediction; (2) CT images are the first time integrated into multimodal fusion for CRC MSI prediction; (3) feature-level fusion strategy is evaluated on both Transformer-based and CNN-based method. Our approach is validated on cross-validation of 352 cases and outperforms either feature-level (0.8177 vs. 0.7908) or decision-level fusion strategy (0.8177 vs. 0.7289) on AUC score.

\end{abstract}

\section{Introduction}

Microsatellite instability (MSI) in colorectal cancer (CRC) determines whether patients with cancer respond exceptionally well to immunotherapy~\cite{sahin2019immune}. Because universal MSI testing requires additional complex genetic or immunohistochemical tests, it is not possible for every patient to be tested for MSI in clinical practice. Therefore, a critical need exists for broadly accessible, cost-efficient tools to aid patient selection for testing. 

Deep learning-based methods have been successfully applied for automated MSI prediction directly from hematoxylin and eosin (H\&E)-stained whole-slide images (WSIs)~\cite{kather2019deep,yamashita2021deep}. Kather et al.~\cite{kather2019deep} developed ResNet-based model to predict patients with MSI and MSS tumors. Another work~\cite{yamashita2021deep} further proposed MSINet and proved the deep learning model exceeded the performance of experienced gastrointestinal pathologists at predicting MSI on WSIs. Despite the vital role of such diagnostic biomarkers~\cite{sidaway2020msi}, patients with similar histology profiles can exhibit diverse outcomes and treatment responses. Novel and more specific biomarkers are needed from a whole spectrum of modalities, ranging from radiology~\cite{pei2022pre,wu2019value,echle2021deep}, histology~\cite{ushizima2022deep,kather2020development,wang2022weakly}, and genomics~\cite{lipkova2022artificial,braman2021deep}.

Given the large complexity of medical data, there are new trends to integrate complementary information from diverse data sources for multimodal data fusion~\cite{chen2022pan,feng2022development,cui2022survival}. Many models have shown the use of radiology data to consider macroscopic factors could achieve more accurate and objective diagnostic and prognostic biomarkers for various cancer types~\cite{wang2019predicting, he2020noninvasive, yao2022deep,dong2020deep}. However, when integrating radiology images and WSIs for predicting MSI, the large data heterogeneity gap between the two modalities exists and makes the integration very difficult. Specifically, a WSI consists of tens of thousands of patches~\cite{chen2021multimodal,lu2021data,wei2019pathologist} while radiology data usually form with 3D shape~\cite{golia2019radiomics}. How to design an effective fusion strategy and learn important interactions between radiology and pathology images is important but still remains unknown for MSI prediction in CRC.  

In this paper, we introduce a new and effective multi-modal fusion pipeline for MSI prediction by combining decision-level fusion and feature-level fusion following Bayesian rules. We also investigated different fusion strategies and found the proposed fusion scheme achieved better results than those methods. The contributions of this paper are: 1) This study generalizes an MSI prediction pipeline in CRC utilizing radiology-guided knowledge.
2) To the best of our knowledge, we are the first to exploit a multi-level fusion strategy for using multi-modal data for MSI prediction.
3) Extensive experimental results suggest the effectiveness of our Bayesian-based multimodal multi-level fusion. It can reduce the gap between pathology and radiology predictions and achieve more robust and accurate fusions than other feature-level or decision-level methods.

\begin{figure}[t]
\begin{center}
\includegraphics[width=1\linewidth]{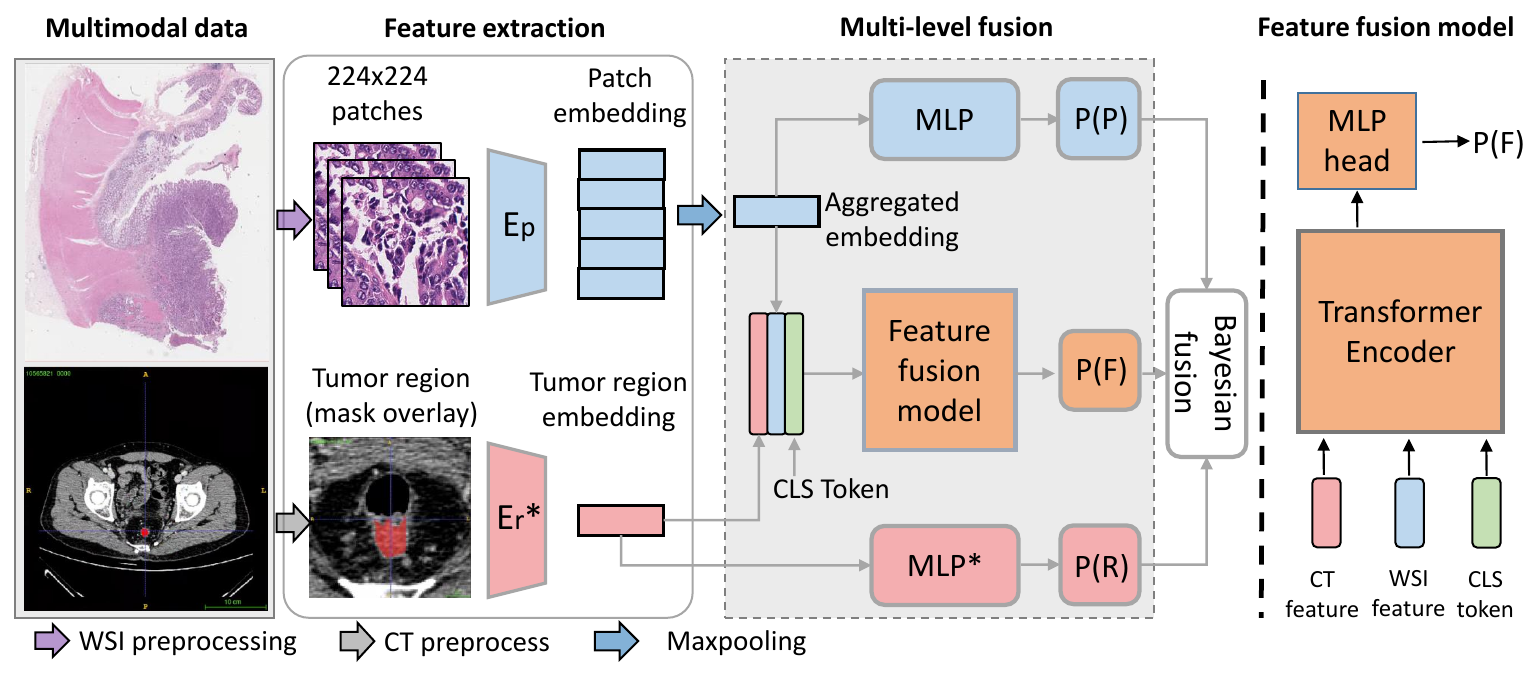}
\end{center}
   \caption{Our proposed $M^{2}$Fusion model. Multimodal data, WSI, and CT images are preprocessed to pathology image patches and CT tumor ROI, respectively. Embeddings are extracted by encoder $E_{p}$ and $E_{r}$. $*$ means the model is well-trained and frozen in pipeline training. $\mathcal{P}_{P}$ is the pathology uni-model performance $\mathcal{P}(P_{ath})$. $P_{R}$ is the radiology uni-model performance $\mathcal{P}(R_{ad})$. $\mathcal{P}_{F}$ is the feature level fusion model probability distribution under pathology and radiology guidance $\mathcal{P}(F_{ea}|P_{ath}R_{ad})$. The final fusion model by $P_{P}$, $P_{R}$ and $P_{F}$ is $\mathcal{P}(F_{ea}P_{ath}R_{ad})$ in Eq.\ref{eq:UVW_p}}
\label{fig:f3}
\end{figure}

\section{Method}

\textbf{Problem Statement.} In our study, each CRC 
patient has a 3D CT image, a pathology whole slide image (WSI), and its corresponding label (MSI status). We aim at CRC MSI prediction using both pathology and radiology data. 
Fig.\ref{fig:f3} shows the proposed Bayesian-based fusion model. 
Our fusion model combines three predictions together and can be seen as feature-level and decision-level fusion in a unified framework.
It consists of two branches that process each modality (pathology or radiology data) and it introduces a radiology feature-guided pathology fusion model. In the following parts, we will discuss why radiology-guided fusion methods could benefit our final prediction.

\subsection{Bayesian-based multi-modality fusion model}
% Denote the information from different modality contains different information. The fusion between two modality images provides extra evidence for the CRC MSI prediction compared with model rely on single modality. Due to the inherit scale difference between the two modality data, 2D gigapixel WSI and 3D CT images, decision level fusion and feature level fusion can provide the interactions from different field of view. 
Assuming the learnable context from each modality is different, we hypothesize that the fusion between modalities knowledge can enhance the confidence level of the CRC MSI prediction, compared with single modality training. Due to the inherent scale difference between the two modalities (2D gigapixel WSI and 3D CT images), we propose a multi-modal fusion strategy, which combines both the decision-level prior and feature-level prior to enhance the interaction between the learnable knowledge from different fields of view. 

%The Bayesian-based model can provide the math proof with Bayesian formula. The general Bayesian formula is Eq.\ref{eq:bayes_basic}.

%\begin{equation} \label{eq:bayes_basic}
%	\mathcal{P}(B|A) = \frac{\mathcal{P}(AB)}{\mathcal{P}(A)}
%\end{equation}

We first define the predictions from pathology data and from radiology data as events $P_{ath}$ and $R_{ad}$, respectively. Here, we hypothesize the probabilistic relationship between prediction with Bayes' theorem as follows:

% Define pathology data predicts MSI status correct as event U. Radiology data predicts MSI status correct as V. We have Eq.\ref{eq:bayes1} according to the Bayesian rule.
\begin{equation} \label{eq:bayes1}
\mathcal{P}(P_{ath}R_{ad})={\mathcal{P}(R_{ad})}\mathcal{P}(P_{ath}|R_{ad}) 
\end{equation}
Here $\mathcal{P}(R_{ad})$ is the uni-model performance on radiology data. $\mathcal{P}(P_{ath}|R_{ad})$ denotes the probabilistic prediction on the model well-trained on pathology data with radiology prior. According to Eq.\ref{eq:bayes1}, if under the guidance of pre-trained radiology model $\mathcal{P}(R_{ad})$, pathology model $\mathcal{P}(P_{ath}|R_{ad})$ performs better than uni-model on pathology ($\mathcal{P}(P_{ath})$), then modality fusion model should perform better than uni-model ($\mathcal{P}(P_{ath})$ and $\mathcal{P}(R_{ad})$).
\begin{equation} \label{eq:UV}
	\mathcal{P}(P_{ath}R_{ad}) \propto  \mathcal{P}(P_{ath}|R_{ad})
\end{equation}

The Bayes' theorem can be extended to three events: feature level multi-modal fusion model predicts MSI status correct as event $F_{ea}$. The extended Bayes' theorem is Eq.\ref{eq:UVW}.
\begin{equation} \label{eq:UVW}
\mathcal{P}(F_{ea}P_{ath}R_{ad})=\mathcal{P}(F_{ea}|P_{ath}R_{ad}) \mathcal{P}(P_{ath}R_{ad})
\end{equation}
Similar to the relation between $\mathcal{P}(P_{ath}|R_{ad})$ and $\mathcal{P}(P_{ath}R_{ad})$, Eq.\ref{eq:UVW_p}. If radiology data can help to get a better feature-level fusion model $\mathcal{P}(F_{ea}|P_{ath}R_{ad})$, the final fusion on both the decision-level and feature-level should outperform the decision-level fusion model.

\begin{equation} \label{eq:UVW_p}
	\mathcal{P}(F_{ea}|P_{ath}R_{ad}) \propto  \mathcal{P}(F_{ea}|P_{ath}R_{ad})
\end{equation}

Bayes' theorem guarantees that if we want to seek a better final fusion model than decision-level fusion, we have to implement a good feature-level fusion model. Our final model could benefit from both feature-level and decision-level fusion.

\subsection{MSI prediction on single modality}
%According to the research mentioned above, CRC MSI prediction only base on single modality data (either CT image or histopathology WSI) achieved promising performance in multiple research works. 
%Here we show each individual model in Fig.~\ref{fig:f1}, the CRC MSI on single modality prediction could be made by two individual path. Path A shows the pipeline for pathology image. Path B introduces the MSI status can also be predicted by radiology data.

%\begin{figure}
%\begin{center}
%\includegraphics[width=0.9\linewidth]{Fig/f1.pdf}
%\end{center}
 %  \caption{Colorectal Cancer MSI prediction on single modality. A. MSI prediction model on pathology Whole Slide Image (WSI), $\mathcal{P}(P)$. B. MSI prediction model on radiology CT image, $\mathcal{P}(R)$. }
%\label{fig:f1}
%\end{figure}

\noindent\textbf{Pathology model.}
Our pathology model is composed of two parts: First, we used the CLAM model\cite{lu2021data} to crop the pathology patches from gigapixel WSI. Second, following the previous work~\cite{yamashita2021deep}, the ResNet-18 is used as an encoder to abstract features from pathology patches. We crop the non-overlapping image tiles in size of $224\times224$ from the WSI foreground. The image patches from all WSI are constructed as a whole pathology patch dataset. The pathology patches label is inherited from the WSI label which it cropped from. The model will predict a patch-level probability of whether the patches belong to MSI or MSS. In the testing phase, the image patches will get the predicted label from the well-trained encoder. The majority vote result of patches from WSI is the patient MSI prediction. \\

\noindent\textbf{Radiology model.}
Based on the 3D radiology CT scans, the tumor region mask of CT volume has been annotated. Two essential slices are cropped from three directions of CT image. One slice is CT tumor region by overlaying the mask on the CT slice. The other slice is the whole CT slice in the direction. The six essential slices (two slices from each direction) are stacked as a six-channel input to build a 2.5D model~\cite{roth2014new}. The encoder used for MSI prediction is ImageNet pre-trained ResNet-18 (modified input channel to six channels). The original 3-channel pre-trained weights are copied to $4^{th}$ to $6^{th}$ channel as initialization.

\begin{figure*}[t]
\begin{center}
\includegraphics[width=1\linewidth]{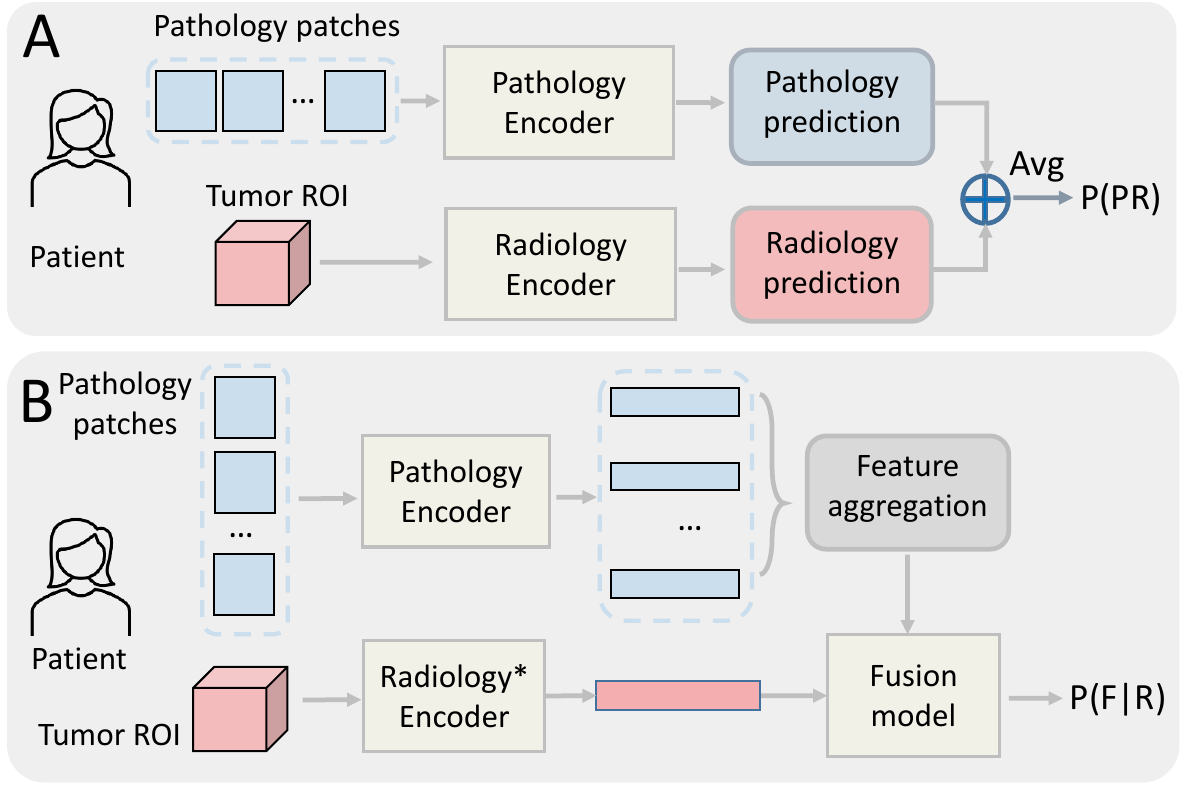}
\end{center}
   \caption{Baseline experiments on multimodal fusion. A. Decision level multimodal fusion, $\mathcal{P}(P_{ath}R_{ad})$ in Eq.\ref{eq:bayes1}. B. Radiology-guided feature-level fusion, probability distribution follows $\mathcal{P}(F_{ea}|R_{ad})$. '*' means the model is well-trained and frozen in pipeline training.}
\label{fig:f2}
\end{figure*}
\subsection{Model prediction fusion on multiple levels}
%To exploit the information from both modality of same patient, we implement multi-modal fusion from two levels: decision level~$\mathcal{P}(PR)$ and feature level~$\mathcal{P}(F|PR)$. 
\subsubsection{Decision level multimodal fusion}
Fig.~\ref{fig:f2}-A shows the decision level fusion. Both models are trained and make the prediction separately. The mean of predicted probability from pathology and radiology is taken as the MSI prediction score for the patient. 
Based on the well-trained uni-model on pathology images and radiology data, the decision-level multimodal fusion employs the patient-level MSI prediction for the final decision. From the well-trained pathology uni-model, the pathology image $W^{i}$ from patient $i$ has predicted MSI probability $P_{p}^{i}$. Similar to pathology prediction, radiology CT scans $C^{i}$ from patient $i$ can get MSi probability prediction $P_{r}^{i}$. The decision level fused prediction follows $P^{i} = (P_{r}^{i} + P_{p}^{i})/2$.

% \begin{equation} \label{eq:decision_fuse}
% 	P^{i} = \frac{(P_{r}^{i} + P_{p}^{i})}{2}
% \end{equation}

\subsubsection{Feature level multi-modal fusion}
Fig.~\ref{fig:f2}-B shows the model fusion on the feature level. The feature embedding abstracted from pathology patches is aggregated as a single feature representing the bag of cropped pathology patches. Each pathology patch is generated as an embedding $e^{i}$ from patch $x^{i}$. The generated embedding $e^{i} \in \mathbb{R}^{1\times512}$ is not representative of the WSI. We first aggregate $e^{i}$ when $i \in [1, N]$ to a single feature for further feature-level fusion. Referring to the Multi-instance Learning (MIL) methods\cite{raju2020graph}, we use maxing pooling on each channel of embeddings to aggregate the single patches embedding to patient pathology embedding $e$. The aggregation process follows Eq.\ref{eq:fea_aggr} where $d \in [0, 511]$ and $e \in \mathbb{R}^{1\times512}$.

\begin{equation} \label{eq:fea_aggr}
	e_{d} = max_{i=0,...,N}e^{i}_{d}
\end{equation}

Radiology feature embedding is abstracted from segmented tumor ROI. The feature embeddings from both modalities are fused by feeding into the fusion model. Two major feature-level fusion strategies are investigated in our study, the Transformer-based or MLP-based fusion model. Transformer model~\cite{dosovitskiy2020image} takes the aggregated WSI feature embedding and radiology ROI embedding as input. Following the standard approach in the transformer model, a learnable class token is added to the input embedding sequence. Multi-layer Perceptron (MLP) fusion model concatenates embeddings from two modalities and is then finetuned with the patient MSI label. The dim of two modality embeddings are both $1\times512$.

\begin{figure*}[h]
\begin{center}
\includegraphics[width=1\linewidth]{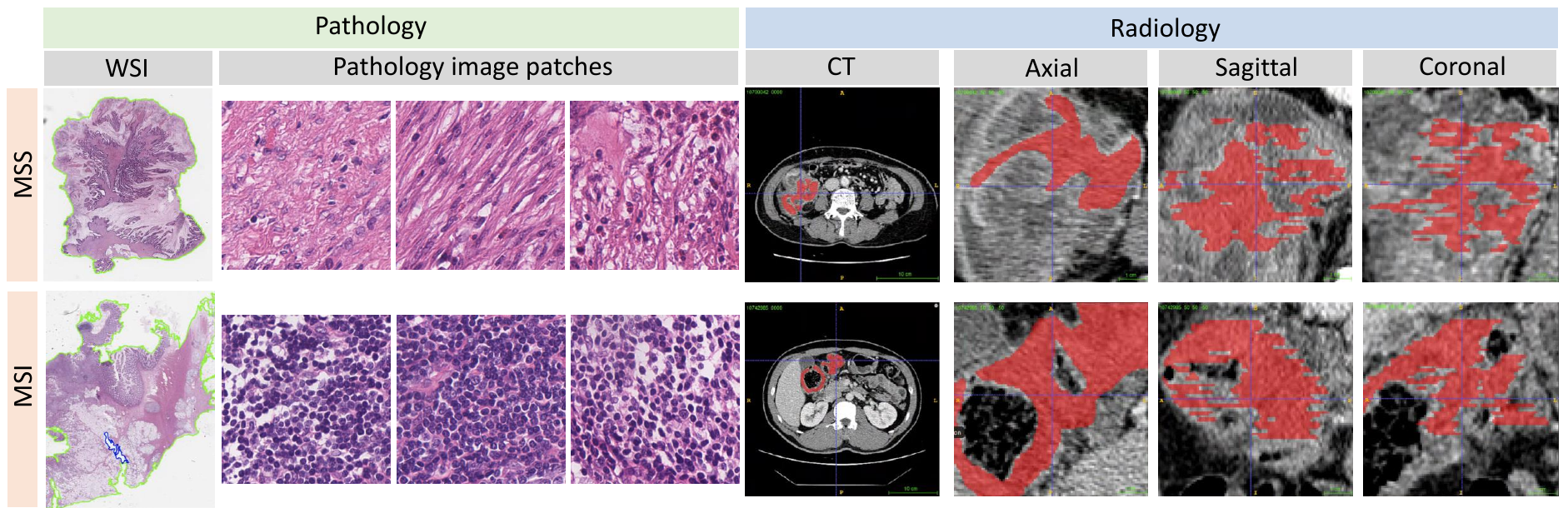}
\end{center}
  \caption{Data visualization of the dataset. First row shows two modalities image from MSS subjects. The second row shows data from MSI subject. }
\label{fig:quality}
\end{figure*}

\section{Experiments}

\subsection{Dataset}
We collect an in-house dataset that has the paired pathology WSIs and CT images from 352 patients shown in Fig.\ref{fig:quality}. The dataset includes 46 MSI patients and 306 MSS patients. The venous phase is used for tumor annotations by a board-certified radiologist with 14 years of specialized experiences. The median imaging spacing is $0.76 \times 0.76 \times 5 $ mm$^3$. The pathology WSI is at a gigapixel level maintained in a pyramid structure. Each level each layer contains a reduced-resolution version of the image from 5$\times$, 10$\times$, and 40$\times$ magnification. The highest level of the pyramid is the full-resolution image which is 40$\times$ in 0.25 $\mu$m per pixel. The image patches are $448\times448$ cropped from 40$\times$ level and resize to $224\times224$. 

To thoroughly evaluate the dataset performance, we use 5-fold cross-validation in all model evaluations. Since the MSI/MSS ratio is unbalanced, the MSI patients and MSS patients are evenly split into five folds to guarantee a fair MSI/MSS ratio in each fold. For each experiment, three folds of data are used for training, one fold for validation, and the rest one fold for testing. By picking up different folds as testing data, five-set experiments are conducted. The average AUC score is used as the evaluation criterion.  

\subsection{Experimental Design}
In the experiments, we aim at evaluating the proposed Bayesian-based multimodal multi-level fusion model. The experiment parts verify two research questions: (1) whether multimodal fusion provides better performance over the uni-model (rely on single data modality), (2) if our proposed Bayesian-based model $\mathcal{P}(F_{ea}P_{ath}R_{ad})$ achieves the optimal fusion strategy over other fusion models. The ablation study is explored feature aggregation and feature-level fusion strategy.

\noindent\textbf{Pathology uni-modal prediction}
The uni-model on pathology data is separated into two steps. First, the WSIs are cropped by the CLAM model into $224\times224$ patches. The patches use the WSI labels in model training. ImageNet-pretrained ResNet-18 is trained for 100 epochs and the batch size is set to 128. In the testing stage, the average probability of patches from the same WSI is used as patient WSI probability prediction. The final model performance is the average score of 5 testing fold.

\noindent\textbf{Radiology uni-modal prediction}
For the Radiology uni-model, we construct the training data by selecting six essential slides based on CT image and annotated tumor region. Only one ROI block is cropped from each CT and constructs the six-channel training data (batch size = 2). ImageNet pre-trained ResNet-18 is employed as the encoder.

\noindent\textbf{Decision level fusion prediction}
Different from uni-model training from scratch, decision-level fusion is based on a well-trained uni-model. Based on the 5-fold well-trained model, we feed the test fold data to the corresponding trained model and get the MSI prediction by pathology data. The same process goes for radiology data. The decision-level fused prediction is computed by average MSI probability from two modalities.

\noindent\textbf{Feature-level fusion prediction} \\
Instead of fusing the probability prediction from two modalities, the regular feature level fusion model fuses the embeddings generated from the two modalities' encoders. Both modality encoders are trained from scratch. 
For the radiology-guided feature level fusion, two modalities of data and a well-trained radiology uni-model are needed. The pathology data is fed into an end-to-end training path. The output of the pathology path is an aggregated feature for pathology WSI. The radiology path is an abstracted feature by pre-trained radiology uni-model from its corresponding training model. For a patient sample, two $1\times512$ features from pathology and radiology data are fed into fusion model. 
For the Transformer-based model, we choose ViT-S as our backbone. Our ViT-S model depth is 8, the head number is 12. Multi-layer perception (MLP) hidden feature dimension is 1024. The input matrix is in $3\times512$. 
CNN-based feature level fusion concatenates the feature from two modalities into one feature with a length of 1024. An MLP is constructed to map the concatenated feature to the final fusion prediction, which has two fully connected layers when the hidden dimension is 256.

% \noindent\textbf{Bayesian-guided feature-level fusion prediction}
% Following the setting of the feature-level fusion model, 

\noindent\textbf{Bayesian-guided multi-level fusion prediction} \\
For the Bayesian-guided fusion model, we used the same input data as previous fusion experiments: a bag of pathology image patches and radiology CT tumor Region of Interest (ROI). The patient MSI prediction from radiology can be generated by the pre-trained model. The feature abstracted from radiology ROI can be generated from the second last layer's output. The feature and patient-level prediction from pathology follow the same procedure as radiology except the pathology encoder is trainable. The fusion model we used is ViT-S for the Transformer-based model and a two-layer MLP for MLP based fusion model. The average score of the pathology, radiology, and feature fusion MSI probability prediction is used as the final prediction.     

\section{Result}
We conduct experiments on 5-fold cross-validation and model performances are shown in Table.~\ref{tab:AUC}. Our proposed multi-level multi-modality fusion pipeline is compared with the single-modality model and fusion methods. From the average AUC score across 5-fold experiments, the performance of unimodal relies on pathology image and radiology image are 0.6847 and 0.7348, respectively. The decision-level fusion has an average AUC score of 0.7908 which outperforms unimodal prediction score. The feature-level fusion model shows better performance by using Vision Transformer than MLP. Without radiology guidance, feature-level fusion model (avg AUC: 0.7289) performs better than pathology unimodal but worse than radiology unimodal.
The radiology data can guide feature-level fusion model training by getting AUC score of 0.7696 better than 0.7289. Radiology-guided feature-level fusion model shows better performance than feature-level fusion without a guide. By combining the decision-level and feature-level information from two image modalities, our proposed multi-level multi-modality pipeline get the best AUC 0.8177 over the rest of MSI CRC strategies.

\begin{table}[h]
\setlength{\tabcolsep}{2.8mm}{
 \caption{AUC on MSI prediction}
  \centering
  \begin{adjustbox}{width=\textwidth}
  \begin{tabular}{l|ccccc|c}
    \toprule
    % \multicolumn{2}{c}{Part}                   \\
    % \cmidrule(r){1-2}
      &  Fold 1  & Fold 2  &  Fold 3  &  Fold 4  & Fold 5  &  Average  \\
    \midrule
    Patho unimodal\cite{yamashita2021deep} & 0.6502  &  0.7282  &  0.8530   &   0.8819   &  0.6500   &  0.6847        \\
    2.5D Radio unimodal & 0.5615  &  0.8333  & 0.7520    &  0.7163    &   0.8158   &  0.7348    \\
    Decision-level fusion &  0.6956 & 0.8313   &  0.8536  &  0.8948   &  0.6785   &  0.7908    \\
       % &  0.5403  & 0.8075   &  0.7937  &  7163   &  0.8063   &  0.7041       \\
    Feature-level fusion & 0.619  &  0.6528   &  0.7698  & 0.7083    &   0.6730   &  0.7289      \\ 
    Radio-guided feature fusion &  0.7218 & 0.7558   &  0.7698  &  0.7678   &  0.8127   &  0.7696   \\ 
    \textbf{$M^{2}$Fusion} & 0.8278 & 0.8055 & 0.7341 & 0.8989 & 0.8222 & \cellcolor{red!25}\textbf{0.8177}\\
    \bottomrule
  \end{tabular}\label{tab:AUC}
  \end{adjustbox}}
\end{table}

% \begin{table}[h]
% \setlength{\tabcolsep}{2.8mm}{
%  \caption{AUC on MSI prediction}
%   \centering
%   \begin{adjustbox}{width=\textwidth}
%   \begin{tabular}{l|ccccc|c}
%     \toprule
%     % \multicolumn{2}{c}{Part}                   \\
%     % \cmidrule(r){1-2}
%       &  Fold 1  & Fold 2  &  Fold 3  &  Fold 4  & Fold 5  &  Average  \\
%     \midrule
%     Patho unimodal & 0.6502  &  0.7282  &  0.8530   &   0.8819   &  0.6500   &  0.6847        \\
%     2.5D Radio unimodal & 0.5615  &  0.8333  & 0.7520    &  0.7163    &   0.8158   &  0.7348    \\
%     Decision-level fusion &  0.6956 & 0.8313   &  0.8536  &  0.8948   &  0.6785   &  0.7908    \\
%        % &  0.5403  & 0.8075   &  0.7937  &  7163   &  0.8063   &  0.7041       \\
%     Feature-level fusion & 0.619  &  0.6528   &  0.7698  & 0.7083    &   0.6730   &  0.7289      \\ 
%     Radio-guided feature fusion &  0.7218 & 0.7558   &  0.7698  &  0.7678   &  0.8127   &  0.7696   \\ 
%     \textbf{$M^{2}$Fusion} & 0.8278 & 0.8055 & 0.7341 & 0.8989 & 0.8222 & \cellcolor{red!25}\textbf{0.8177}\\
%     \bottomrule
%   \end{tabular}\label{tab:AUC}
%   \end{adjustbox}}
% \end{table}

An ablation study on exploring the pathology feature aggregation strategy and multimodal feature level fusion backbone is shown in Table.\ref{tab:AUC_abla}. The combination of average pooling on pathology feature aggregation and using a Transformer as feature-level fusion backbone has the best AUC performance. 

\begin{table}[h]
\setlength{\tabcolsep}{2.8mm}{
 \caption{Ablation study for pathology feature aggregation and feature-level fusion strategy}
  \centering
  \begin{adjustbox}{width=\textwidth}
  \begin{tabular}{cc|ccccc|c}
    \toprule
    % \multicolumn{2}{c}{Part}                   \\
    % \cmidrule(r){1-2}
     Feature aggregation  &   Feature fusion    &  Fold 1  & Fold 2  &  Fold 3  &  Fold 4  & Fold 5  &  Average  \\
    \midrule
    
     Conv  &  Transformer  &  0.5423 & 0.6012   &  0.7976  &  0.7540   &  0.7746   &  0.6939   \\ 
     Avg  &  CNN  &  0.5786 & 0.7004   &  0.7202  &  0.7044   &  0.7333   &  0.6874   \\ 
     Conv  &  CNN &  0.6593 & 0.7321   &  0.7599  &  0.6706   &  0.7047   &  0.7053   \\ 
     Avg  &  Transformer   & 0.7218  &  0.7758   &  0.7698  & 0.7678    &   0.8127   &  \cellcolor{red!25}\textbf{0.7696}   \\
    % \textbf{Ours} & 0.8278 & 0.8055 & 0.7341 & 0.8989 & 0.8222 & \textbf{0.8177}\\
    \bottomrule
  \end{tabular}  \label{tab:AUC_abla}
  \end{adjustbox}}
\end{table}

% The qualitative result is shown in Fig.~\ref{fig:quality}. Two patients are selected to show our multi-modal fusion strategy can aggregate the information from two modalities to make more accurate predictions. The first row is MSS patients. Pathology unimodal prediction has a large gap with radiology prediction. Decision-level fusion model makes a 0.5504 prediction probability which is closer to the clinical annotation compared with pathology prediction. Our model shows a more accurate prediction than the decision fusion model. For second-row MSI patient case, evidence is clear in the CT image which has 0.6029 model prediction probability when pathology prediction is far from clinical annotation. Our proposed fusion model has a 0.5598 prediction score by utilizing both modalities' information.

\section{Conclusion}
We proposed a multi-level multi-modality fusion pipeline for colorectal cancer MSI status prediction based on pathology WSIs and CT images. We introduce Bayes' theorem to fuse the information from two image modalities on both the feature level and decision level. The experiment result shows (1) radiology and pathology image fusion (decision level fusion) helps CRC MSI prediction by combining the two modalities' information from the same patient, and (2) radiology-guided feature-level training outperforms the model that directly fuses two modalities' features. Our Bayesian-based fusion on both decision-level and feature-level achieves the best performance.

% \section{Acknowledgement}
% 

% \textbf{Acknowledgements}:
% *****

% \clearpage
\bibliographystyle{splncs04}
\bibliography{main}




% LNCS Homepage, \url{http://www.springer.com/lncs}. Last accessed 4
% \end{thebibliography}
\end{document}